\documentclass{article}
\usepackage{amsmath}
\PassOptionsToPackage{numbers,compress}{natbib}

\usepackage[preprint]{neurips_2026}

\usepackage[utf8]{inputenc} 
\usepackage[T1]{fontenc}    
\usepackage{hyperref}       
\usepackage{url}            
\usepackage{booktabs}       
\usepackage{amsfonts}       
\usepackage{nicefrac}       
\usepackage{microtype}      
\usepackage{xcolor}         
\usepackage{graphicx}
\usepackage{multirow}
\usepackage{amssymb}    
\usepackage{array}
\usepackage{placeins}
\usepackage[most]{tcolorbox}
\newtcolorbox{promptbox}{
  breakable,
  colback=white,
  colframe=black!35,
  boxrule=0.4pt,
  arc=1.5pt,
  left=6pt,
  right=6pt,
  top=5pt,
  bottom=5pt,
  boxsep=0pt
}
\newcolumntype{C}[1]{>{\centering\arraybackslash}p{#1}}

\title{VL-SAM-v3: Memory-Guided Visual Priors for Open-World Object Detection}

\author{Chih-Chung Liu \quad
Zhiwei Lin \quad
Yongtao Wang\thanks{Corresponding author} \\
Wangxuan Institute of Computer Technology, Peking University, China \\
\texttt{\{zzliu25\}@stu.pku.edu.cn} \quad
\texttt{\{zwlin, wyt\}@pku.edu.cn}
}

\begin{document}

\maketitle

\begin{abstract}
Open-world object detection aims to localize and recognize objects beyond a fixed closed-set label space. 
It is commonly divided into two categories, \textit{i.e.}, open-vocabulary detection, which assumes a predefined category list at test time, and open-ended detection, which requires generating candidate categories during the inference. 
%
Existing methods rely primarily on coarse textual semantics and parametric knowledge, which often provide insufficient visual evidence for fine-grained appearance variation, rare categories, and cluttered scenes. 
In this paper, we propose VL-SAM-v3, a unified framework that augments open-world detection with retrieval-grounded external visual memory. Specifically, once candidate categories are available, VL-SAM-v3 retrieves relevant visual prototypes from a non-parametric memory bank and transforms them into two complementary visual priors, \textit{i.e.}, sparse priors for instance-level spatial anchoring and dense priors for class-aware local context. 
These priors are integrated with the original detection prompts via Memory-Guided Prompt Refinement, enabling a shared retrieval-and-refinement mechanism that supports open-vocabulary and open-ended inference.
%
Extensive zero-shot experiments on LVIS show that VL-SAM-v3 consistently improves detection performance under both open-vocabulary and open-ended inference, with particularly strong gains on rare categories.
Moreover, experiments with a stronger open-vocabulary detector (\textit{i.e.}, SAM3) validate the generality of the proposed retrieval-and-refinement mechanism.
\end{abstract}
\section{Introduction}

Open-world object detection \citep{joseph2021towards,gupta2022ow} aims to localize and recognize objects beyond a fixed set of predefined categories. Unlike conventional closed-set detectors \citep{ren2015faster,redmon2016you}, which rely on exhaustive instance-level supervision over a fixed label space, open-world detection offers a more scalable route toward visual understanding. 
In practice, it is commonly clustered into two types, \textit{i.e.,} \emph{open-vocabulary detection}, which assumes a predefined category list at test time, and \emph{open-ended detection}, which requires candidate categories during the inference. Despite this difference in candidate acquisition, both settings ultimately depend on semantic descriptions to guide detection.

Recent advances in vision-language learning have made both settings increasingly practical \citep{peng2023openscene}. Open-vocabulary detectors \citep{li2022grounded,kim2024retrieval} can generalize beyond training labels by conditioning on category text, while open-ended systems \citep{lin2024generative,lin2024training} further remove the need for a predefined category list.
However, both paradigms remain constrained by coarse textual semantics and parametric knowledge. Language can specify \emph{what} to detect, but often provides only limited clues about \emph{how} the target may appear in the image. This limitation is particularly severe for long-tail categories, cluttered scenes, and objects with substantial intra-class variation \citep{chen2024multi,schrodi2024two}. For example, a text prompt such as ``dog'' compresses large variations in breed, pose, viewpoint, and surrounding context into a single concept, making it difficult for the detector to recover the fine-grained visual cues needed for accurate localization and discrimination.


This observation suggests that open-world detectors need more than text alone. 
Extra visual evidence is required for more accurate detection. 
Recently, retrieval-augmented models have shown that non-parametric memory can complement parametric knowledge by recovering fine-grained evidence at inference time \citep{lewis2020retrieval}.
However, current research on how to retrieval essential visual evidence for open-world detection is limited.

To address this challenge, we propose {VL-SAM-v3}, a unified open-world detection framework that augments detector prompting with external visual memory. Given an input image and candidate categories, VL-SAM-v3 retrieves relevant visual prototypes from a non-parametric memory bank and converts them into two complementary priors: \emph{sparse priors}, which provide instance-level spatial anchors, and \emph{dense priors}, which provide class-aware local visual context. These retrieval-grounded priors are then integrated with original detector prompts through a \emph{Memory-Guided Prompt Refinement} module. The resulting retrieval-and-refinement mechanism applies to both open-vocabulary and open-ended inference, yielding a unified architecture for the two settings.

\begin{figure}[t]
\centering
\includegraphics[width=\textwidth]{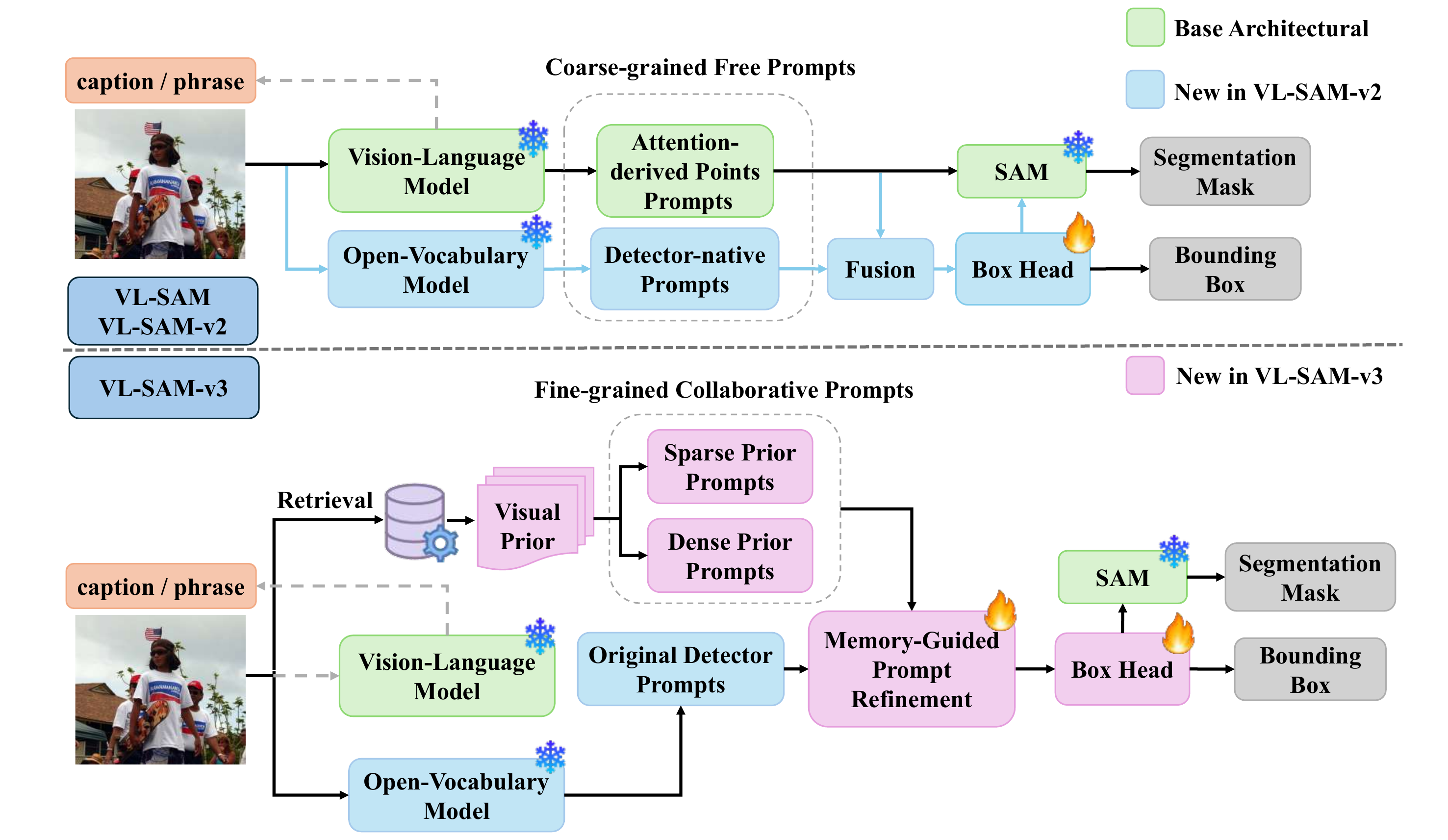}
\caption{\textbf{Comparison of VL-SAM/VL-SAM-v2 and VL-SAM-v3.} VL-SAM-v3 extends previous variants with retrieval-grounded visual priors and Memory-Guided Prompt Refinement.}
\label{fig:pipeline}
\end{figure}

The main contributions of this work are summarized as follows:
\begin{itemize}
\item We propose VL-SAM-v3, a unified framework for open-world detection that supports both open-vocabulary and open-ended inference by augmenting detector prompts with external visual memory.
\item We introduce a retrieval mechanism and Memory-Guided Prompt Refinement to transform non-parametric visual evidence into sparse and dense priors, and fuse with original prompts.
\item We demonstrate that VL-SAM-v3 consistently improves zero-shot open-vocabulary and open-ended detection on LVIS, especially for rare categories.
\end{itemize}
\section{Related work}

\subsection{Open-world detection and prompt adaptation}

Open-world detection extends object recognition beyond a fixed closed-set label space \citep{joseph2021towards,gupta2022ow}. A common setting is \emph{open-vocabulary detection}, in which a predefined category list is provided at inference time and the detector predicts from this list. Representative methods such as GLIP \citep{li2022grounded} and Grounding-DINO \citep{liu2024grounding} strengthen vision-language fusion for phrase grounding and detection, while later works further improve vocabulary coverage, efficiency, and semantic reasoning \citep{cheng2024yolo,yao2024detclipv3,fu2025llmdet}. Despite this progress, open-vocabulary detectors still require explicit category candidates at test time.
More recent studies have explored \emph{open-ended detection}, in which candidate semantics are generated online by a vision-language model rather than specified in advance \citep{lin2024generative,lin2024training}. This setting offers greater flexibility, but its performance depends more critically on the quality of the generated categories, which may be ambiguous, incomplete, or noisy. Furthermore, open-vocabulary and open-ended detection share a common limitation, \textit{i.e.}, they rely heavily on coarse textual semantics and often struggle to capture fine-grained appearance variation, long-tail categories, and cluttered visual contexts \citep{chen2024multi}.

Prompt adaptation has recently been introduced to better align pretrained vision-language representations with downstream detection tasks \citep{zhou2022learning}. Methods such as DetPro \citep{du2022learning}, PromptDet \citep{feng2022promptdet}, and MQ-Det \citep{xu2023multi} show that improved prompt construction or visual exemplars can enhance category retrieval and detection.
However, these methods mainly focus on open-vocabulary detection.


In contrast, this paper proposes VL-SAM-v3, a unified framework that augments prompt-based detection with retrieval-grounded visual priors for open-vocabulary and open-ended detection.

\subsection{Retrieval-augmented models and external memory}

Retrieval-augmented models complement parametric knowledge with non-parametric memory \citep{lewis2020retrieval,gao2023retrieval}. By retrieving relevant evidence at inference time, these systems can improve factuality, robustness, and adaptability beyond what is stored in model parameters alone.

In vision, external memory and retrieval have been explored for tasks such as classification and captioning \citep{long2022retrieval,li2024evcap}, where retrieved examples provide category-specific or context-specific evidence. Extending retrieval to dense prediction is more challenging, because the retrieved evidence must support both semantic recognition and spatial grounding. Our method addresses this challenge by converting retrieved visual evidence into sparse and dense priors for open-world detection.
\section{Method}
\label{sec:method}

\begin{figure*}[t]
\centering
\includegraphics[width=\textwidth]{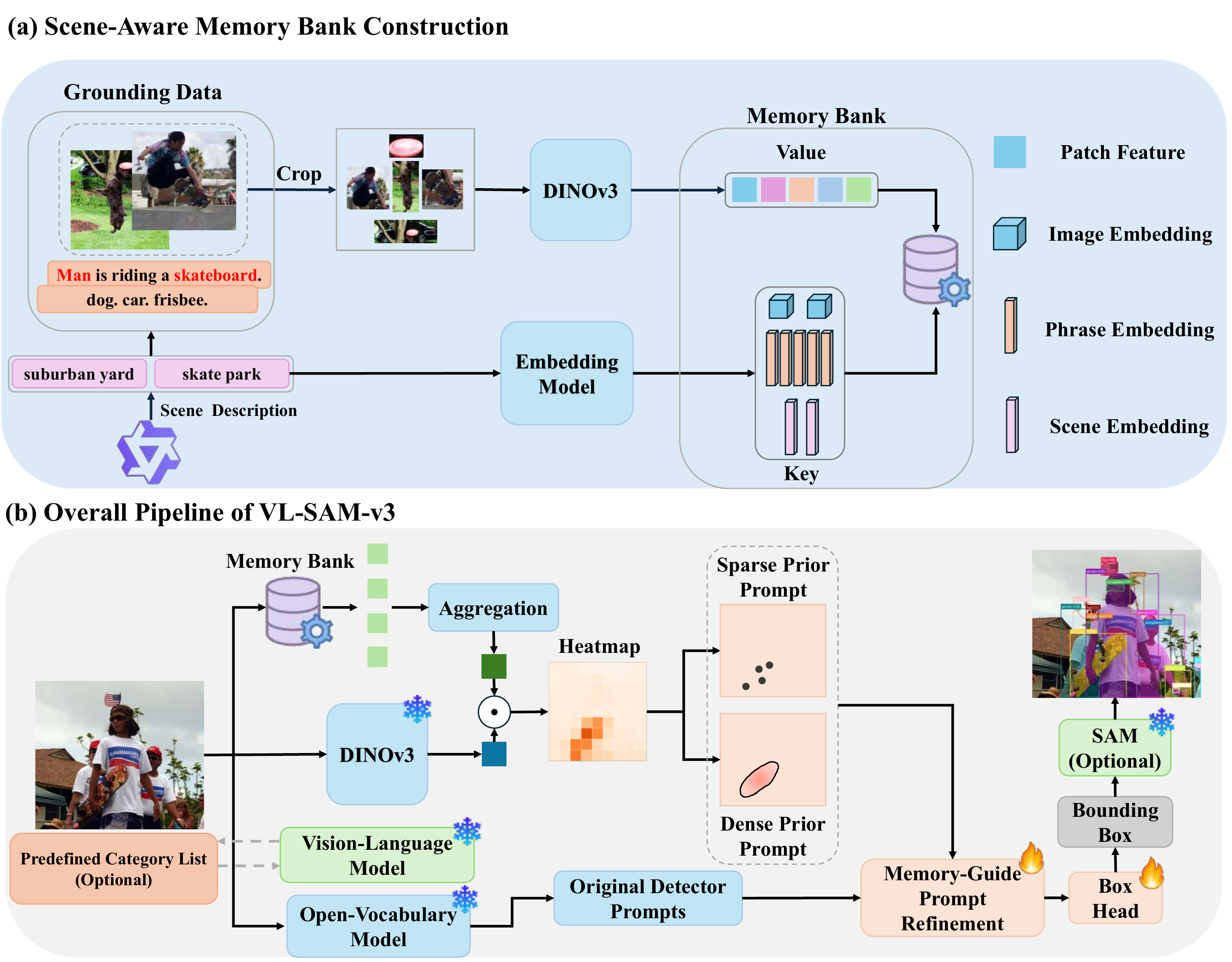}
\caption{\textbf{Overview of VL-SAM-v3.} (a) Scene-aware visual memory construction from grounding-style data. (b) Retrieval-augmented open-world detection, where retrieved visual evidence is converted into sparse and dense visual priors and injected into the detector through Memory-Guided Prompt Refinement.}
\label{fig:method_pipeline}
\end{figure*}

Figure~\ref{fig:method_pipeline} provides an overview of VL-SAM-v3.
Given an image $I$ and a candidate category set $\mathcal{C}$, we augment a prompt-based open-world detector with retrieval-grounded external visual memory.
For each category, we retrieve relevant visual evidence from a scene-aware memory bank, convert it into sparse and dense visual priors on the current image, and inject these priors into the detector through {Memory-Guided Prompt Refinement}.

\subsection{Problem setup}

Given an image $I$, the goal is to localize and recognize objects from a candidate category set $\mathcal{C}$.
In open-vocabulary detection, $\mathcal{C}$ is provided at test time, in open-ended, it is generated during the inference.
VL-SAM-v3 uses the same procedure in both settings and differs only in how $\mathcal{C}$ is obtained.


\subsection{Scene-aware visual memory}
\label{sec:Scene-aware visual memory}


We build a scene-aware visual memory from grounding-style data, where each entry stores a retrieval key $k_i$ and a region-level visual value $v_i$ for a grounded phrase-region pair.
Specifically, given image $I$, the ground-truth bounding box $b_i$, and the corresponding category or phrase $t_i$, we first use VLM to describe the scene $s$ of image $I$.
Then, we calculate $k_i$ and $v_i$ as:
\[
k_i
=
\operatorname{Norm}\!\big(
w_p E_{\mathrm{text}}(t_i)
+
w_s E_{\mathrm{text}}(s)
+
w_g E_{\mathrm{img}}(I)
\big),
\]
\[
v_i
=
\operatorname{Norm}\!\big(
\operatorname{Pool}(F(I), b_i)
\big),
\]
where $w_p$, $w_s$, and $w_g$ are weights for phrase content, scene context, and image-level visual context, respectively.
$E_{\mathrm{text}}$ and $E_{\mathrm{img}}$ denote the text and image encoders of the multimodal embedding model.
$F(\cdot)$ represents the DINOv3 feature encoder~\citep{simeoni2025dinov3}.
$\operatorname{Norm}(\cdot)$ is $\ell_2$ normalization, and $\operatorname{Pool}(F, b_i)$ denotes mean pooling over the DINOv3 patch features inside box $b_i$.
In this equation, $k_i$ is optimized for scene-aware retrieval, while $v_i$ preserves the appearance of a grounded region in the DINOv3 feature space.

\subsection{Category-conditioned retrieval}

For each candidate category $c\in\mathcal{C}$, we form a query $q_c$ in the same space as the memory keys $k_i$:
\[
q_c
=
\operatorname{Norm}\!\big(
w_p E_{\mathrm{text}}(c)
+
w_s E_{\mathrm{text}}(s)
+
w_g E_{\mathrm{img}}(I)
\big).
\]
This query combines category text with the scene and global visual context of the current image.

Then, we calculate the similarity between $q_c$ and each $k_i$ in visual memory, and select the top-$K$ retrieved entries and their indices $\mathcal{N}_c$.
Finally, the retrieved values are then aggregated into a category-specific visual prototype $p_c$ as the output:
\[
p_c
=
\operatorname{Norm}\!\left(
\sum_{i\in\mathcal{N}_c}\alpha_i\, v_i
\right),
\]
where $\alpha_i$ is calculated as:
\[
\alpha_i
=
\frac{\exp(\langle q_c, k_i\rangle / \tau_p)}
{\sum_{m\in\mathcal{N}_c}\exp(\langle q_c, k_m\rangle / \tau_p)}.
\]
The prototype $p_c$ summarizes the retrieved evidence for category $c$ in the DINOv3 feature space.


\subsection{Retrieval-grounded visual priors}

We next project $p_c$ onto the input image $I_{input}$ to derive two complementary visual priors: a dense prior for class-aware spatial support and a sparse prior for instance-level anchors.

\paragraph{Dense prior.}
We compute a category-specific heatmap as:
\[
H_c
=
\operatorname{MinMax}\!\Big(
\operatorname{Smooth}\!\big(
\langle \operatorname{Norm}(F(I_{\mathrm{input}})),\, p_c \rangle
\big)
\Big),
\]
where $\operatorname{Smooth}(\cdot)$ denotes spatial smoothing and $\operatorname{MinMax}(\cdot)$ rescales the heatmap to $[0,1]$.
$H_c$ measures the compatibility between the input image feature and the retrieved prototype $p_c$.

\paragraph{Sparse prior.}
Because $H_c$ provides dense support but not explicit instance centers, we extract sparse anchors from its local maxima. Specifically, we first sort candidate peaks by response and greedily retain spatially diverse ones using distance-based suppression, yielding $\mathcal{A}_c=\{a_j\}$.
Then, $H_c$ and $\mathcal{A}_c$ form the retrieval-grounded visual priors.

\subsection{Memory-Guided Prompt Refinement}

\begin{figure*}[t]
\centering
\includegraphics[width=\textwidth]{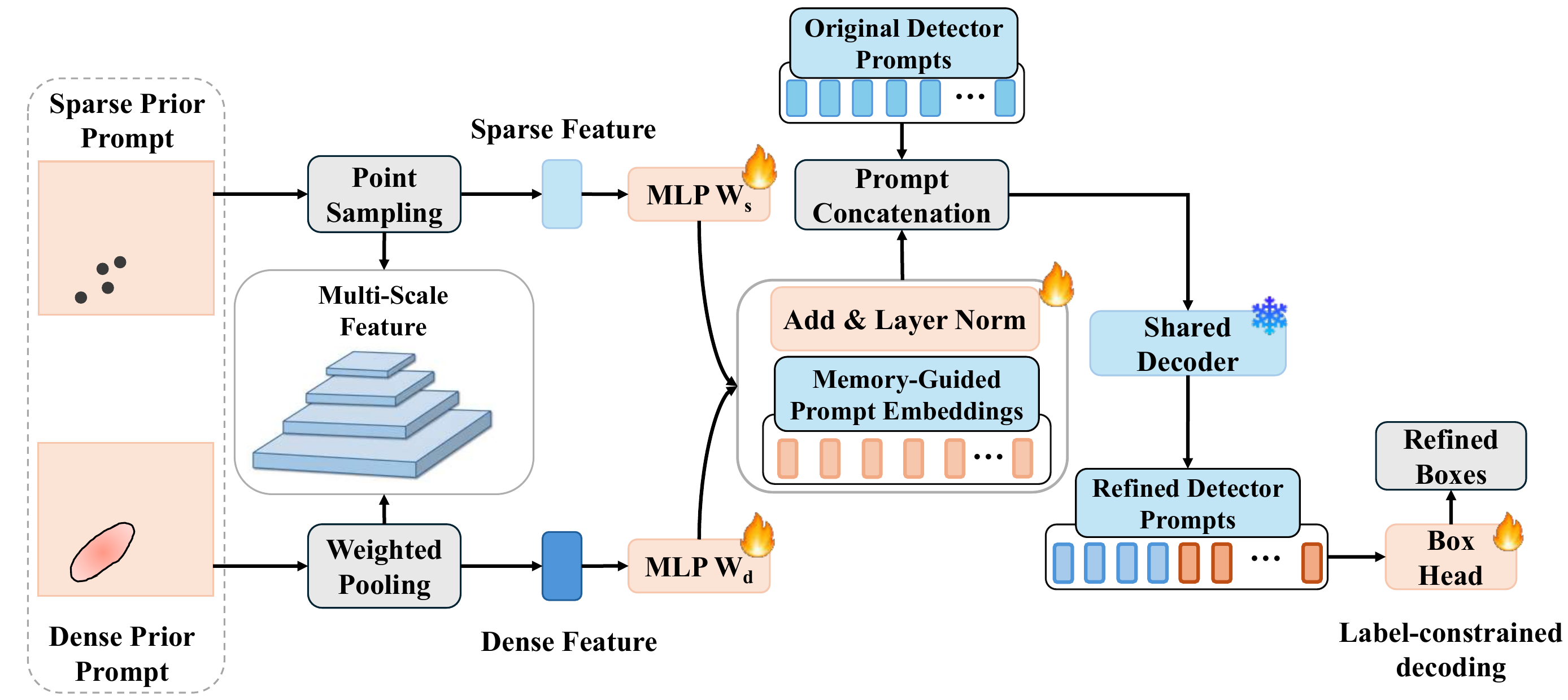}
\caption{\textbf{Illustration of Memory-Guided Prompt Refinement.} Sparse anchors and dense priors produce sparse and dense features, which are fused into memory-guided prompt embeddings and decoded together with the detector's original prompts.}
\label{fig:memory_guided_refinement}
\end{figure*}

Given the dense prior $H_c$ and sparse anchors $\mathcal{A}_c$, we construct memory-guided prompt embeddings for category $c$, as shown in Figure~\ref{fig:memory_guided_refinement}.
%
Specifically, for the original detector's multi-scale features, we perform the following refinement independently at each scale.
Let $M$ and $e$ denote the detector's features and the original learnable prompt prior.
%
We first extract a sparse feature for each anchor $a_j\in\mathcal{A}_c$ from the detector's feature :
\[
f^s_j
=
\operatorname{Sample}(M, a_j),
\]
where $\operatorname{Sample}(\cdot,\cdot)$ denotes bilinear feature sampling at the anchor location.


After that, to extract a dense feature from $\mathcal{A}_c$, we construct local windows $\Omega_j$ centered at each anchor $a_j$.
Within $\Omega_j$, we utilize $H_c$ as the weights to compute the dense feature from $M$ as:
\[
f^d_j
=
\sum_{(x,y)\in\Omega_j} H_c^{x,y}\, M^{x,y}.
\]

Finally, we fuse the sparse feature $f^s_j$, the dense feature $f^d_j$, and the detector's original prompt prior $e$ as:
\[
z_j
=
\operatorname{LN}\!\left(
e
+
W_s f^s_j
+
W_d f^d_j
\right),
\]
where $W_s$ and $W_d$ are learnable projections and $\operatorname{LN}(\cdot)$ is layer normalization.
Each refined prompt combines the detector's original query prior with two memory-derived cues, \textit{i.e.}, a sparse instance-level cue and a dense local-context cue.

This procedure is applied to each feature scale in the detector, and the refined prompts from all scales are concatenated with the detector's original prompts and decoded by the same transformer decoder.


\subsection{Label-constrained decoding}

After refinement across all encoder levels, we collect the resulting memory-guided prompts $\{z_j\}$.
Each memory-guided prompt $z_j$ is associated with its category $t_j$ in Section \ref{sec:Scene-aware visual memory}.
To prevent cross-category drift, the memory-guided prompt is allowed to vote only for its source category.
Let $s_j(c)$ denote the classification logit of prompt $z_j$ for input candidate category $c \in \mathcal{C}$.
We enforce the constraint by:
\[
\tilde{s}_j(c)
=
\begin{cases}
s_j(c), & c = t_j,\\
-\infty, & c \neq t_j.
\end{cases}
\]

That is, prompt $z_j$ contributes only to category $t_j$, while the detector's original prompts remain unconstrained for global open-world search.

\paragraph{Training and inference.}
During training, we avoid online retrieval by using ground-truth box centers as anchor approximations and precomputing class-specific dense priors offline with the same memory mechanism.
We then optimize the detector with the same supervision as the base detector.
During inference, retrieval is performed online for each candidate category to construct $p_c$, $H_c$, $\mathcal{A}_c$, and the corresponding memory-guided prompts.
\section{Experiments}

\subsection{Experimental setup}
\label{sec:setup}

\paragraph{Benchmarks and protocols.}
We primarily evaluate VL-SAM-v3 on LVIS~\citep{gupta2019lvis} under the zero-shot setting and report fixed AP together with AP on rare, common, and frequent categories, denoted as AP$_r$, AP$_c$, and AP$_f$~\citep{dave2021evaluating}. 
We evaluate both open-vocabulary and open-ended inference. For open-ended detection, Table~\ref{tab:main_oe} explicitly reports the base detector and candidate generator for each method. Rows with the same base detector and candidate generator provide controlled comparisons that isolate detector-side improvements, while rows with stronger candidate generators examine the complementarity between category generation and retrieval-grounded refinement.
Following LLMDet~\citep{fu2025llmdet}, we also report COCO~\citep{lin2014microsoft} results for reference, although these are not zero-shot because GroundingCap-1M contains COCO images.

\paragraph{Implementation details.}
We instantiate VL-SAM-v3 with two base open-vocabulary detectors. 
Our main implementation builds on LLMDet~\citep{fu2025llmdet}, and initializes the Swin-T and Swin-L variants from the corresponding pretrained checkpoints. 
We follow LLMDet for the training strategy and loss functions, and fine-tune the detector on GroundingCap-1M~\citep{fu2025llmdet} for 150k iterations with a total batch size of 16 on 8 NVIDIA A100 GPUs.

To evaluate the generality of the proposed retrieval-and-refinement mechanism, we further apply VL-SAM-v3 to SAM3~\citep{carion2025sam}, a state-of-the-art open-vocabulary detector. 
For a controlled comparison, all SAM3 experiments use the same fine-tuning data and evaluation protocols as our main setting, with fine-tuning on GroundingCap-1M for 2 epochs and evaluation on LVIS under open-vocabulary and open-ended protocols.

The external memory bank is constructed offline from the grounding annotations of GroundingCap-1M. 
Scene descriptors are generated by Qwen3-VL-8B~\citep{bai2025qwen3}, and the retrieval keys are computed by Qwen3-VL-Embedding-2B~\citep{li2026qwen3}. Region-level visual prototypes are extracted from DINOv3~\citep{simeoni2025dinov3}. 
To avoid contamination, we exclude all images overlapping with the LVIS evaluation splits when building the memory bank, and during training we further exclude entries from the current image to prevent trivial self-matching. 
Unless otherwise specified, we retrieve the top-$K$ entries with $K=12$, and set $w_p=1.0$, $w_s=0.3$, and $w_g=0.01$. For aggregation, we set $\tau_p=0.07$. Additional implementation details are provided in Appendix~\ref{app:implementation}, and inference cost is deferred to Appendix~\ref{app:inference_cost}.

\subsection{Main results}
\begin{table}[!t]
\centering
\caption{\textbf{Zero-shot open-vocabulary object detection results on LVIS \textit{minival} and \textit{val}.}}
\label{tab:main_ov}
\footnotesize
\setlength{\tabcolsep}{3.2pt}
\renewcommand{\arraystretch}{1.08}
\begin{tabular}{lccccccccc}
\toprule
\multicolumn{1}{l}{\multirow{2}{*}{Method}} &
\multicolumn{1}{c}{\multirow{2}{*}{Backbone}} &
\multicolumn{4}{c}{LVIS minival} &
\multicolumn{4}{c}{LVIS val} \\
\cmidrule(lr){3-6} \cmidrule(lr){7-10}
& & AP & AP$_r$ & AP$_c$ & AP$_f$ & AP & AP$_r$ & AP$_c$ & AP$_f$ \\
\midrule
GLIP~\citep{li2022grounded} & Swin-T & 26.0 & 20.8 & 21.4 & 31.0 & 17.2 & 10.1 & 12.5 & 25.2 \\
GLIPv2~\citep{zhang2022glipv2} & Swin-T & 29.0 & -- & -- & -- & -- & -- & -- & -- \\
CapDet~\citep{long2023capdet} & Swin-T & 33.8 & 29.6 & 32.8 & 35.5 & -- & -- & -- & -- \\
Grounding-DINO~\citep{liu2024grounding} & Swin-T & 27.4 & 18.1 & 23.3 & 32.7 & 20.1 & 10.1 & 15.3 & 29.9 \\
OWL-ST~\citep{minderer2023scaling} & CLIP B/16 & 34.4 & 38.3 & -- & -- & 28.6 & 30.3 & -- & -- \\
Desco-GLIP~\citep{li2023desco} & Swin-T & 34.6 & 30.8 & 30.5 & 39.0 & 26.2 & 19.6 & 22.0 & 33.6 \\
DetCLIP~\citep{yao2022detclip} & Swin-T & 35.9 & 33.2 & 35.7 & 36.4 & 28.4 & 25.0 & 27.0 & 28.4 \\
DetCLIPv2~\citep{yao2023detclipv2} & Swin-T & 40.4 & 36.0 & 41.7 & 40.4 & 32.8 & 31.0 & 31.7 & 34.8 \\
YOLO-World-L~\citep{cheng2024yolo} & YOLOv8-L & 35.4 & 27.6 & 34.1 & 38.0 & -- & -- & -- & -- \\
T-Rex2~\citep{jiang2024t} & Swin-T & 42.8 & 37.4 & 39.7 & 46.5 & 34.8 & 29.0 & 31.5 & 41.2 \\
OV-DINO~\citep{wang2024ov} & Swin-T & 40.1 & 34.5 & 39.5 & 41.5 & 32.9 & 29.1 & 30.4 & 37.4 \\
LLMDet~\citep{fu2025llmdet} & Swin-T & 44.7 & 37.3 & 39.5 & 50.7 & 34.9 & 26.0 & 30.1 & 44.3 \\
VL-SAM-v2~\citep{lin2025vl} & Swin-T & 45.7 & 41.2 & 41.1 & 50.7 & 35.5 & 29.3 & 31.8 & 44.3 \\
\textbf{VL-SAM-v3 (Ours)} & \textbf{Swin-T} & \textbf{47.7} & \textbf{44.9} & \textbf{44.9} & \textbf{50.7} & \textbf{38.1} & \textbf{35.2} & \textbf{33.7} & \textbf{44.3} \\
\midrule
GLIP~\citep{li2022grounded} & Swin-L & 37.3 & 28.2 & 34.3 & 41.5 & 26.9 & 17.1 & 23.3 & 36.4 \\
GLIPv2~\citep{zhang2022glipv2} & Swin-H & 50.1 & -- & -- & -- & -- & -- & -- & -- \\
Grounding-DINO~\citep{liu2024grounding} & Swin-L & 33.9 & 22.2 & 30.7 & 38.8 & -- & -- & -- & -- \\
OWL-ST~\citep{minderer2023scaling} & CLIP L/14 & 40.9 & 41.5 & -- & -- & 35.2 & 36.2 & -- & -- \\
DetCLIP~\citep{yao2022detclip} & Swin-L & 38.6 & 36.0 & 38.3 & 39.3 & 28.4 & 25.0 & 27.0 & 31.6 \\
DetCLIPv2~\citep{yao2023detclipv2} & Swin-L & 44.7 & 43.1 & 46.3 & 43.7 & 36.6 & 33.3 & 36.2 & 38.5 \\
DetCLIPv3~\citep{yao2024detclipv3} & Swin-L & 48.8 & 49.9 & 49.7 & 47.8 & 41.4 & 41.4 & 40.5 & 42.3 \\
LLMDet~\citep{fu2025llmdet} & Swin-L & 51.1 & 45.1 & 46.1 & 56.6 & 42.0 & 31.6 & 38.8 & 50.2 \\
VL-SAM-v2~\citep{lin2025vl} & Swin-L & 51.7 & 47.2 & 46.7 & 56.6 & 42.5 & 33.2 & 39.7 & 50.2 \\
\textbf{VL-SAM-v3 (Ours)} & \textbf{Swin-L} & \textbf{53.4} & \textbf{50.9} & \textbf{50.8} & \textbf{56.6} & \textbf{43.5} & \textbf{39.7} & \textbf{38.3} & \textbf{50.2} \\
\midrule
SAM3~\citep{carion2025sam} & PE-L+ & 59.1 & 61.9 & 59.0 & 58.8 & 53.6 & 54.9 & 51.1 & 55.1 \\
\textbf{VL-SAM-v3 + SAM3} & \textbf{PE-L+} & \textbf{60.2} & \textbf{63.4} & \textbf{59.9} & \textbf{58.8} & \textbf{54.1} & \textbf{55.8} & \textbf{51.7} & \textbf{55.1} \\
\bottomrule
\end{tabular}
\end{table}
\paragraph{Open-vocabulary detection on LVIS.}
Table~\ref{tab:main_ov} reports zero-shot open-vocabulary detection results on LVIS \textit{minival} and \textit{val}. 
For the LLMDet-based instantiations, VL-SAM-v3 consistently improves over the corresponding baselines with both backbones. With Swin-T, it reaches 47.7 AP on \textit{minival} and 38.1 AP on \textit{val}, and with Swin-L, it further reaches 53.4 AP and 43.5 AP on \textit{minival} and \textit{val}, respectively. 
Furthermore, the gains are concentrated on rare and common categories. On \textit{minival}, VL-SAM-v3 improves AP$_r$/AP$_c$ from 37.3/39.5 to 44.9/44.9 with Swin-T, and from 45.1/46.1 to 50.9/50.8 with Swin-L. Similar trends hold on \textit{val}, while AP$_f$ remains largely unchanged under the same inference protocol as VL-SAM-v2~\citep{lin2025vl}. 
These results are consistent with our motivation: retrieval-grounded visual evidence is most useful when coarse text semantics and parametric detector knowledge are insufficient, especially for long-tail and visually diverse categories.

The SAM3 results in Table~\ref{tab:main_ov} show the same trend: VL-SAM-v3 further improves over the strong SAM3 baseline on both \textit{minival} and \textit{val}, indicating that the proposed retrieval-and-refinement mechanism is not tied to LLMDet.

\paragraph{Open-ended detection on LVIS.}
Table~\ref{tab:main_oe} reports zero-shot open-ended detection results on LVIS \textit{minival}, where candidate categories must be generated during inference rather than provided in advance. 
This setting is more challenging because errors in category generation directly affect downstream grounding.

Controlled comparisons under the same candidate generator isolate the detector-side contribution of VL-SAM-v3. Using the same InternVL-2.5 (8B) candidate generator as VL-SAM-v2, VL-SAM-v3 improves AP from 29.5 to 40.6 with Swin-T and from 31.8 to 41.7 with Swin-L. The corresponding AP$_r$ gains are from 29.8 to 37.9 and from 30.5 to 38.7. 
These results show that retrieval-grounded prompt refinement substantially improves grounding quality even when the category candidates are held fixed.
Furthermore, replacing InternVL-2.5 with the stronger Qwen3-VL (8B) generator improves VL-SAM-v3 to 41.4 AP with Swin-T and 44.8 AP with Swin-L. 
This suggests that stronger category generation and retrieval-grounded refinement are complementary: better candidate categories help, while the controlled comparisons show that a substantial part of the improvement comes from the detector-side retrieval-and-refinement mechanism itself.

The SAM3-based comparison further supports this conclusion: under the same Qwen3-VL generator, VL-SAM-v3 + SAM3 improves over SAM3 from 49.9/52.6 to 51.7/54.0 AP/AP$_r$.
\begin{table}[t]
\centering
\caption{\textbf{Zero-shot open-ended object detection results on LVIS \textit{minival}.}}
\label{tab:main_oe}
\small
\setlength{\tabcolsep}{3.6pt}
\begin{tabular}{lccc}
\toprule
Method & Base detector / backbone & Candidate generator & AP / AP$_r$ \\
\midrule
GenerateU~\citep{lin2024generative} & Swin-T & FlanT5-base & 26.8 / 20.0 \\
GenerateU~\citep{lin2024generative} & Swin-L & FlanT5-base & 27.9 / 22.3 \\
Open-Det~\citep{cao2025open} & Swin-L & FlanT5-base & 33.1 / 31.2 \\
VL-SAM~\citep{lin2024training} & ViT-H & CogVLM (17B) & 25.3 / 23.4 \\
\midrule
VL-SAM-v2~\citep{lin2025vl} & LLMDet / Swin-T & InternVL-2.5 (8B) & 29.5 / 29.8 \\
VL-SAM-v3 & LLMDet / Swin-T & InternVL-2.5 (8B) & 40.6 / 37.9 \\
VL-SAM-v3 & LLMDet / Swin-T & Qwen3-VL (8B) & 41.4 / 39.4 \\
\midrule
VL-SAM-v2~\citep{lin2025vl} & LLMDet / Swin-L & InternVL-2.5 (8B) & 31.8 / 30.5 \\
VL-SAM-v3 & LLMDet / Swin-L & InternVL-2.5 (8B) & 41.7 / 38.7 \\
VL-SAM-v3 & LLMDet / Swin-L & Qwen3-VL (8B) & 44.8 / 41.9 \\
\midrule
SAM3~\citep{carion2025sam} & SAM3 / PE-L+ & Qwen3-VL (8B) & 49.9 / 52.6 \\
\textbf{VL-SAM-v3 + SAM3} & \textbf{SAM3 / PE-L+} & \textbf{Qwen3-VL (8B)} & \textbf{51.7 / 54.0} \\
\bottomrule
\end{tabular}
\end{table}

Additional transfer results to open-ended instance segmentation are reported in Appendix~\ref{app:open_ended_instance_segmentation}.

\paragraph{Reference results on COCO.}
Table~\ref{tab:coco} reports COCO AP following LLMDet~\citep{fu2025llmdet}. 
Because GroundingCap-1M includes COCO images, these results do not constitute a zero-shot evaluation and are reported only for reference. 
Under this setting, VL-SAM-v3 achieves higher AP than both LLMDet and VL-SAM-v2.

\begin{table}[t]
\centering
\caption{\textbf{Open-vocabulary object detection results on COCO.} Gray AP values indicate that COCO images are included in the training dataset.}
\label{tab:coco}
\small
\setlength{\tabcolsep}{14pt}
\begin{tabular}{lcc}
\toprule
Method & Backbone & COCO AP \\
\midrule
GLIP~\citep{li2022grounded} & Swin-T & 46.1 \\
DetCLIPv3~\citep{yao2024detclipv3} & Swin-T & 47.2 \\
Grounding-DINO~\citep{liu2024grounding} & Swin-T & 48.4 \\
MM-GDINO~\citep{zhao2024open} & Swin-T & 50.4 \\
\midrule
LLMDet~\citep{fu2025llmdet} & Swin-T & \textcolor{gray}{55.6} \\
VL-SAM-v2~\citep{lin2025vl} & Swin-T & \textcolor{gray}{56.0} \\
\textbf{VL-SAM-v3 (Ours)} & \textbf{Swin-T} & \textcolor{gray}{\textbf{56.8}} \\
\bottomrule
\end{tabular}
\end{table}

\subsection{Ablation and analysis}

We next analyze the key design choices of VL-SAM-v3 on LVIS \textit{minival} with the Swin-T backbone. More ablations on the external visual memory are provided in Appendix~\ref{app:ablation_memory}.
\begin{table}[!htbp]
\centering
\caption{\textbf{Ablation of sparse and dense priors.}}
\label{tab:ablation_prior}
\small
\setlength{\tabcolsep}{6pt}
\begin{tabular}{cccc}
\toprule
Sparse prior & Dense prior & AP & AP$_r$ \\
\midrule
 &  & 44.7 & 37.3 \\
\checkmark &  & 46.8 & 42.1 \\
 & \checkmark & 46.7 & 41.6 \\
\checkmark & \checkmark & \textbf{47.7} & \textbf{44.9} \\
\bottomrule
\end{tabular}
\end{table}
\paragraph{Effect of retrieval-grounded priors.}

Table~\ref{tab:ablation_prior} ablates the sparse and dense priors derived from external memory. 
Starting from the no-memory baseline (44.7 AP, 37.3 AP$_r$), adding the sparse prior improves performance to 46.8 AP and 42.1 AP$_r$, while adding the dense prior yields 46.7 AP and 41.6 AP$_r$. 
Combining them gives the best result, reaching 47.7 AP and 44.9 AP$_r$. 
These results indicate that the sparse and dense priors are complementary: the sparse prior provides instance-level spatial anchoring, while the dense prior provides class-aware local support.


\paragraph{Effect of scene descriptors and label-constrained decoding.}
Table~\ref{tab:ablation_scene_las} evaluates two lightweight components. 
Adding scene descriptors improves AP/AP$_r$ from 47.1/43.1 to 47.7/44.9, showing that scene context helps retrieve more relevant visual evidence than category phrases alone. 
Applying label-constrained decoding improves AP/AP$_r$ from 46.9/42.7 to 47.7/44.9, indicating that binding each memory-guided prompt to its source category reduces cross-category drift during decoding.

\begin{table}[!htbp]
\centering
\caption{\textbf{Ablation of scene descriptors and label-constrained decoding.}}
\label{tab:ablation_scene_las}
\small
\begin{minipage}{0.47\linewidth}
\centering
\begin{tabular}{ccc}
\toprule
Scene descriptor & AP & AP$_r$ \\
\midrule
 & 47.1 & 43.1 \\
\checkmark & \textbf{47.7} & \textbf{44.9} \\
\bottomrule
\end{tabular}
\end{minipage}
\hfill
\begin{minipage}{0.47\linewidth}
\centering
\begin{tabular}{ccc}
\toprule
Label-constrained decoding & AP & AP$_r$ \\
\midrule
 & 46.9 & 42.7 \\
\checkmark & \textbf{47.7} & \textbf{44.9} \\
\bottomrule
\end{tabular}
\end{minipage}
\end{table}

\paragraph{Robustness to auxiliary vision-language models.}
Table~\ref{tab:ablation_vlm} evaluates VL-SAM-v3 with different auxiliary vision-language models, including GLM4.6V-Flash~\citep{hong2025glm45v}, InternVL-2.5~\citep{chen2024expanding}, InternVL-3.5~\citep{wang2025internvl3}, and Qwen3-VL~\citep{bai2025qwen3}, under the same open-vocabulary protocol with all other settings fixed. 
VL-SAM-v3 performs consistently well across all tested models, while stronger models generally lead to slightly better results.
These results indicate that the proposed retrieval-and-refinement mechanism does not depend on a particular auxiliary vision-language model.

\begin{table}[htbp]
\centering
\caption{\textbf{Ablation of auxiliary vision-language models.}}
\label{tab:ablation_vlm}
\small
\setlength{\tabcolsep}{5.5pt}
\begin{tabular}{lccc}
\toprule
Method & Auxiliary model & AP & AP$_r$ \\
\midrule
LLMDet~\citep{fu2025llmdet} & -- & 44.7 & 37.3 \\
\midrule
\multirow{4}{*}{VL-SAM-v3}
& GLM4.6V-Flash (9B) & 47.0 & 41.3 \\
& InternVL-2.5 (8B) & 46.8 & 41.0 \\
& InternVL-3.5 (8B) & 47.1 & 43.8 \\
& Qwen3-VL (8B) & \textbf{47.7} & \textbf{44.9} \\
\bottomrule
\end{tabular}
\end{table}

\FloatBarrier
\section{Conclusion}
\label{sec:conclusion}

We presented VL-SAM-v3, an open-world object detection framework that augments prompt-based detection with retrieval-grounded external visual memory. By retrieving fine-grained visual prototypes and converting them into sparse and dense priors, VL-SAM-v3 injects detector-compatible visual evidence beyond coarse textual semantics. Through Memory-Guided Prompt Refinement, the framework provides a shared retrieval-and-refinement mechanism for both open-vocabulary and open-ended inference. Extensive experiments on LVIS show consistent gains in both settings, with particularly significant improvements in open-ended settings. 
These results highlight the value of retrieval-grounded visual memory for open-world detection.

\noindent\textbf{Broader Impacts Statement.}
This paper investigates retrieval-guided visual memory for open-world object detection, which may benefit automatic labeling systems and open-world visual perception systems. We do not identify major privacy-related risks, but future deployment should consider potential biases from external visual memory and vision-language models.
\newpage
\FloatBarrier
\bibliographystyle{plainnat}
\bibliography{main}

\newpage
\appendix
\section{Additional implementation details}
\label{app:implementation}

This section provides additional implementation details for the external visual memory in VL-SAM-v3. We first describe how scene descriptors are generated, and then summarize the lightweight filtering strategy used during memory construction and retrieval.

\subsection{Scene descriptor generation}
\label{app:scene_descriptor}

For each image, we generate a single coarse scene descriptor using Qwen3-VL-8B~\citep{bai2025qwen3}. The goal is to capture the dominant environmental context of the image rather than object-level semantics. The descriptor is obtained with the following prompt:

\medskip
\begin{promptbox}
\small
\textbf{Prompt.} \emph{Identify the single scene category that best describes the overall environment of this image. Focus on the global setting or location type (e.g., ``kitchen'', ``forest'', ``beach'', ``restaurant dining area'', ``subway station'') rather than individual objects, people, attributes, or actions. Output exactly one concise scene label and nothing else.}
\end{promptbox}
\medskip

We set the decoding parameters to \texttt{max\_new\_tokens=128}, \texttt{do\_sample=true}, \texttt{temperature=0.7}, \texttt{top\_p=0.8}, \texttt{top\_k=20}, and \texttt{repetition\_penalty=1.0}. Restricting the model to output a single scene label helps prevent the descriptor from drifting toward object-level descriptions, which would otherwise partially overlap with the category query itself.

\subsection{Filtering during memory construction and retrieval}
\label{app:filtering}

We apply lightweight filtering both when building the memory bank and when retrieving entries at inference time.

\paragraph{Construction-time filtering.}
During memory construction, we first remove extremely small boxes, since such regions are often too limited to provide reliable visual features. We then apply a blur-based quality filter. Specifically, duplicated boxes are merged, and the remaining unique regions are scored using the variance of the Laplacian. We discard the lowest-quality 10\% of regions according to this score, which helps suppress visually ambiguous or severely degraded memory entries.

\paragraph{Retrieval-time filtering.}
At inference time, FAISS~\citep{douze2025faiss} first recalls 200 candidate entries for each query. When exact rescoring is enabled, these candidates are re-ranked using exact similarity scores before selecting the final top-$K$ results. This two-stage procedure preserves the efficiency of approximate nearest-neighbor search while reducing retrieval errors caused by coarse ANN ranking.

\section{Additional ablations on external visual memory}
\label{app:ablation_memory}

This section provides additional ablations for the design choices of the external visual memory. We analyze both the representation used for memory values and the number of retrieved entries used to construct the category-specific prototype.

\subsection{Value representation of memory entries}
\label{app:value_representation}

Each memory entry stores a semantic key together with a region-level visual value. For the value representation, we compare two alternatives: the CLS token extracted from the cropped region and the mean-pooled patch embeddings of the region. Table~\ref{tab:app_value_repr} reports the results on LVIS \textit{minival}~\citep{gupta2019lvis} using the Swin-T backbone~\citep{liu2021swin}.

\begin{table}[h]
\centering
\caption{Effect of the value representation used in the memory bank on LVIS \textit{minival}.}
\label{tab:app_value_repr}
\small
\begin{tabular}{lcc}
\toprule
Value representation & AP & AP$_r$ \\
\midrule
CLS token & 46.4 & 42.7 \\
Patch-embedding mean & \textbf{47.7} & \textbf{44.9} \\
\bottomrule
\end{tabular}
\end{table}

Mean-pooled patch embeddings outperform the CLS-token representation by 1.3 AP and 2.2 AP$_r$. We attribute this gain to the stronger locality of patch features, which better preserve region-level appearance details and are therefore more helpful for rare or fine-grained categories. Based on this result, we use mean-pooled patch embeddings as the default memory value throughout the paper.

\subsection{Effect of retrieval top-$K$}
\label{app:topk}

Table~\ref{tab:app_topk} studies the number of retrieved memory entries used to construct the category-specific visual prototype. All experiments are conducted on LVIS \textit{minival} with the Swin-T backbone.

\begin{table}[h]
\centering
\caption{Effect of retrieval top-$K$ on LVIS \textit{minival}.}
\label{tab:app_topk}
\small
\begin{tabular}{ccc}
\toprule
Top-$K$ & AP & AP$_r$ \\
\midrule
4  & 47.0 & 43.3 \\
8  & 47.4 & 44.2 \\
12 & \textbf{47.7} & \textbf{44.9} \\
16 & 47.3 & 44.4 \\
\bottomrule
\end{tabular}
\end{table}

Performance improves as $K$ increases from 4 to 12, suggesting that multiple retrieved prototypes provide more stable and diverse visual evidence than only a few nearest neighbors. Increasing $K$ further to 16 leads to a slight performance drop, likely because lower-ranked retrieved entries introduce additional noise. We therefore adopt $K=12$ as the default setting.

\section{Additional experimental results}
\label{app:additional_experimental_results}

\subsection{Transfer to open-ended instance segmentation}
\label{app:open_ended_instance_segmentation}

Following Grounded-SAM~\citep{ren2024grounded}, we further combine VL-SAM-v3 with SAM~\citep{kirillov2023segment} for open-ended instance segmentation. 
SAM is used only as an optional post-processing module that converts refined detection boxes into masks, without changing detector training.  
As shown in Table~\ref{tab:main_mask}, VL-SAM-v3 reaches 39.9 mask AP and 36.4 mask AP$_r$ on LVIS \textit{minival}. 
Although the compared methods use different candidate generators and are therefore not directly matched, the large gain suggests that the detector-side improvements of VL-SAM-v3 transfer favorably to mask prediction.

\begin{table}[!htbp]
\centering
\caption{\textbf{Zero-shot open-ended instance segmentation results on LVIS \textit{minival}.} We report mask AP and mask AP$_r$.}
\label{tab:main_mask}
\small
\setlength{\tabcolsep}{6pt}
\begin{tabular}{lccc}
\toprule
Method & Candidate generator & mask AP & mask AP$_r$ \\
\midrule
VL-SAM~\citep{lin2024training} & CogVLM (17B) & 23.9 & 22.7 \\
VL-SAM-v2~\citep{lin2025vl} & InternVL-2.5 (8B) & 28.7 & 27.7 \\
\textbf{VL-SAM-v3} & \textbf{Qwen3-VL (8B)} & \textbf{39.9} & \textbf{36.4} \\
\bottomrule
\end{tabular}
\end{table}

\section{Efficiency and scalability}
\label{app:efficiency}

This section summarizes the storage, retrieval, and runtime cost of VL-SAM-v3. We first report the scale and retrieval efficiency of the memory bank, and then present the end-to-end inference cost in the open-ended setting.

\subsection{Memory bank scale and retrieval cost}
\label{app:memory_scale}

Table~\ref{tab:app_memory_scale} summarizes the size of the memory bank and the corresponding retrieval cost. For efficient approximate search, we use a FAISS IndexIVFPQ index~\citep{jegou2010product} with inner-product similarity.

\begin{table}[h]
\centering
\caption{Memory bank scale and retrieval cost.}
\label{tab:app_memory_scale}
\small
\begin{tabular}{lc}
\toprule
Item & Value \\
\midrule
Number of entries & 10.83M \\
FAISS index type & IndexIVFPQ \\
Similarity metric & Inner product \\
Total storage & 63.97 GiB \\
Storage per entry & 6.20 KiB \\
FAISS retrieval speed & $\sim$400 queries/s \\
\bottomrule
\end{tabular}
\end{table}

The reported retrieval speed measures only the FAISS search stage and does not include query encoding or the detector forward pass. These statistics show that the proposed memory bank remains practical at multi-million-entry scale while preserving sufficiently fast approximate retrieval.

\subsection{Inference cost}
\label{app:inference_cost}

Table~\ref{tab:inference_cost} reports the inference cost of VL-SAM-v3 in the open-ended setting on a single NVIDIA A100 GPU with 80G memory using Swin-T and Qwen3-VL (8B). The detector-side overhead of memory-guided refinement is lightweight: sparse and dense prompt generation takes 80 ms per image, while the refinement head adds only 5 ms. The dominant cost comes from vision-language generation rather than the retrieval-and-refinement module itself.

\begin{table}[h]
\centering
\caption{Inference cost of VL-SAM-v3 in the open-ended setting.}
\label{tab:inference_cost}
\small
\setlength{\tabcolsep}{7pt}
\begin{tabular}{lcc}
\toprule
Component & Time / image (ms) & GPU memory (G) \\
\midrule
Scene descriptor generation (optional) & 500 & $\sim$20 \\
Category generation & 700 & $\sim$22 \\
Sparse \& dense prompt generation & 80 & $\sim$7 \\
Refinement \& head & 5 & $\sim$2 \\
\bottomrule
\end{tabular}
\end{table}



\end{document}